\documentclass[journal]{IEEEtran}
\usepackage{amsmath}
\usepackage{amssymb}
\usepackage{graphicx}
\usepackage{caption}
\usepackage{cite}
\usepackage{epstopdf}
\usepackage{url}
\DeclareMathOperator*{\argmax}{arg\,max}

\begin{document}
\title{The Impact of Quantity of Training Data on Recognition of Eating Gestures}

\author{Yiru Shen,
            Eric Muth,
            and Adam Hoover~\IEEEmembership{Senior Member, IEEE}
\thanks{Y. Shen and A. Hoover are with the Department of Electrical
and Computer Engineering, Clemson University, Clemson, SC 29634-0915 USA
(e-mail:  yirus@clemson.edu; ahoover@clemson.edu).}
\thanks{E. Muth is with the Department of Psychology,
Clemson University, Clemson, SC 29634 USA
(e-mail: muth@clemson.edu).}
}

\maketitle

\begin{abstract}

This paper considers the problem of recognizing eating gestures by tracking wrist motion.
Eating gestures can have large variability in motion depending on the subject, utensil, and type of food or beverage being consumed.
Previous works have shown viable proofs-of-concept of recognizing eating gestures in laboratory settings with small numbers of subjects and food types, but it is unclear how well these methods would work if tested on a larger population in natural settings.
As more subjects, locations and foods are tested, a larger amount of motion variability could cause a decrease in recognition accuracy.
To explore this issue, this paper describes the collection and annotation of 51,614 eating gestures taken by 269 subjects eating a meal in a cafeteria.
Experiments are described that explore the complexity of hidden Markov models (HMMs) and the amount of training data needed to adequately capture the motion variability across this large data set.
Results found that HMMs needed a complexity of 13 states and 5 Gaussians to reach a plateau in accuracy, signifying that a minimum of 65 samples per gesture type are needed.
Results also found that 500 training samples per gesture type were needed to identify the point of diminishing returns in recognition accuracy.
Overall, the findings provide evidence that the size a data set typically used to demonstrate a laboratory proofs-of-concept may not be sufficiently large enough to capture all the motion variability that could be expected in transitioning to deployment with a larger population.
Our data set, which is 1-2 orders of magnitude larger than all data sets tested in previous works, is being made publicly available.

\end{abstract}

\begin{IEEEkeywords}
activity recognition, gesture recognition, hidden Markov models (HMM)
\end{IEEEkeywords}

\IEEEpeerreviewmaketitle

\section{Introduction}
\label{intro}

\IEEEPARstart{T}{his} 
paper is motivated by recent advances in body-worn sensors for automatic monitoring of energy intake \cite{sazonov2010energetics, lara2013survey, fontana2015detection}.
Wrist-worn wearable devices containing accelerometers and gyroscopes can be used to recognize eating related gestures \cite{amft2009body, dong2012new, junker2008gesture, Shen17}.
Gesture recognition has been widely studied in the domain of
sign language recognition \cite{fang2007large, zaki2011sign}, motivating a similar approach for eating gesture recognition \cite{Ramos15}. 
However, the variability in motion of an eating gesture is much larger than the variability in motion of a sign language gesture.
Sign language gestures are specifically designed to communicate intent, and subject training is conducted to minimize variability in repeated execution of the same gesture.
In contrast, eating gestures are a result of a physiological activity (eating) and their execution varies depending on many variables including the subject, utensil, and food or beverage consumed.
This paper explores the necessary complexity for a hidden Markov model (HMM) to adequately capture the motion variability in eating gestures.
We also test the effect of the quantity of training data needed to adequately train the HMMs.

The idea that a data set must be large enough to represent the behaviors of the population of interest is a common problem in the social sciences.
A recent large project attempted to replicate 21 studies published in Nature and Science and found that with a five fold increase in the size of the populations tested, only 13 studies achieved similar results, and the effect sizes decreased by about 50\% on average compared with the original works \cite{camerer2018evaluating}.
With respect to automatically sensing and measuring eating behaviors, a classifier needs to capture between subject differences (different people may move differently while eating the same foods), within subject differences (the same person may behave differently over multiple meals), and situational differences (eating in the home vs a restaurant, eating alone vs with others, etc).
Therefore the construction of a classifier requires a lot of data with ground truth across a range of subjects, over a period of time, and in a variety of situations.  
The challenge is that it is tremendously difficult to capture the ground truth of the microstructure of eating behaviors in free living.
This paper makes some progress towards that goal, specifically on the axis of a large number of subjects.

For sign language recognition, it is expected that a small amount of training data and model complexity are sufficient. 
One study found that an HMM constructed with 3 states and 3 Gaussians achieved 92\% accuracy in recognizing 5,113 different words \cite{fang2007large}. 
Another study found that 4 states with only 1 Gaussian achieved 91\% accuracy in recognizing 25 different words \cite{kumar2017coupled}. 
For training data, one study found that 24 training samples per word achieved 95\% accuracy for differentiating 300 different words \cite{mohandes2012signer}. 
Another study found that 60 training samples per word achieved 91\% accuracy for differentiating 30 different words \cite{al2009video}. 
The general approach in all these works is to vary model complexity and/or quantity of training data to identify the point of diminishing returns in recognition accuracy. 
This paper describes a similar effort for the recognition of eating gestures.
The question is whether or not a similar model complexity (3-4 states, 1-3 Gaussians) and amount of training data (24-60 samples per word) are sufficient.

For eating gesture recognition, one study found that an HMM constructed with 5 states and 1 Gaussian achieved 94\% in recognizing 384 gestures from 2 subjects \cite{amft2005detection}.
Another study found that an HMM constructed with 13 states and 5 Gaussians achieved 84.3\% in recognizing 2,786 gestures from 25 subjects \cite{Ramos15}.
For training data, one study found that 760 intake gestures from 10 subjects could train models that achieved 93\% accuracy \cite{ye2015automatic}.
Another study found that 1,184 intake gestures from 14 subjects could train models that achieved up to 76\% F1 score \cite{thomaz2017exploring}.
While the accuracies reported in these studies provide evidence of viable proofs-of-concept, it is unclear what accuracies would be achieved if the same methods were deployed on a larger population outside the laboratory.

The novelty of this paper is as follows.  
First, we collected a data set of eating gestures 1-2 orders of magnitude larger than all previous works studying this approach (see Table \ref{table:statitics-gestures}).  
Second, we investigate the effect of model complexity on recognition accuracy on this large data set.  
Third, we investigate the effect of the quantity of training data on recognition accuracy.
The overall goal of this paper is to provide context on the relationship between a laboratory experiment that demonstrates a proof-of-concept, and a potential deployment of the same method on a larger population.
Because of the inherent variability in eating gestures, one would expect that a classifier or algorithm trained on a small amount of laboratory data would not necessarily achieve the same accuracy when tested on a larger population.
The experiments in this paper provide some evidence on the size of training data and model complexity needed to provide confidence in that translation.

\begin{table}
\centering
\begin{tabular}{| c | c | c |} \hline 
Study & \#Subjects & \#Gestures \\ \hline
\cite{Ramos15} & 25 & 2,786 \\ 
\cite{amft2005detection} & 2 & 384 \\
\cite{ye2015automatic} & 10 & 760 \\ 
\cite{thomaz2017exploring} & 14 & 1184 \\
This work & 269 & 51,614 \\ \hline
\end{tabular}
\caption{Statistics of data set in eating gestures.}
\label{table:statitics-gestures}
\end{table}

\section{Methods}
\label{sec:methods}

\subsection{Data Collection}
\label{subsec:data}

\begin{figure}
\begin{center}
\begin{tabular}{c c}
\includegraphics[height=1.3in]{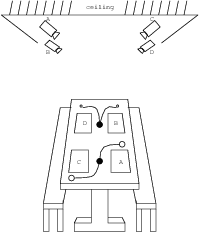} &
\includegraphics[height=1.3in]{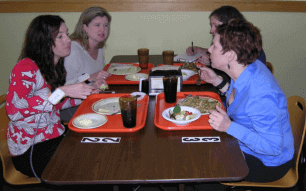} 
\end{tabular}
\end{center}
\caption{The table instrumented for data collection.
Each participant wore a custom tethered device to track wrist motion.}
\label{fig:eating_table}
\end{figure}

Data recording took place in the Harcombe Dining Hall at Clemson University. 
Figure \ref{fig:eating_table} shows an illustration of an instrumented table that can record data up to four subjects simultaneously \cite{Huang13}. 
Four digital video cameras in the ceiling (approximately 5 meters height)
were used to record each participant’s mouth, torso, and tray during
meal consumption. 
A custom device was designed to record the wrist motion of subjects at 15 Hz during eating, using MEMS accelerometers (STMicroelectronics LIS344ALH) to measure x, y and z axis accelerations, and gyroscopes (STMicroelectronics LPR410AL) to measure yaw, pitch and roll rotational velocities.
Cameras and wrist motion trackers were wired to the same computers and used timestamps for synchronization.
A total of 276 subjects were recruited and each consumed a single meal \cite{Shen17}. 
For 5 subjects, either the video or wrist motion tracking failed to record, and for 2 subjects non-dominant hands were used for recording; these are excluded from analysis.
Demographics of the subjects are 129 male, 140 female; age 18-75; height 50-77 in (127-195 cm); weight 100-335 lb (45-152 kg); self-identified ethnicity 26 African American, 28 Asian or Pacific Islander, 189 Caucasian, 11 Hispanic, 15 Other.

Five gesture types were defined based on participant intent during eating as observed via video \cite{Ramos15}.
A bite gesture indicates the movements of taking a bite of food,
a drink gesture indicates the movements of taking a drink of liquid,
a utensiling gesture indicates the movements of manipulating food for
preparation of intake,
and a rest gesture indicates not moving.
Any other activity such as using a napkin or gesturing while talking is
designated as an other gesture.
On average, segmenting and labeling the sensor data of each meal took
3-5 hours and in total the process took more than 1,000 man-hours of work.

\begin{figure*}
	\centering
	\includegraphics[width=0.75\textwidth]{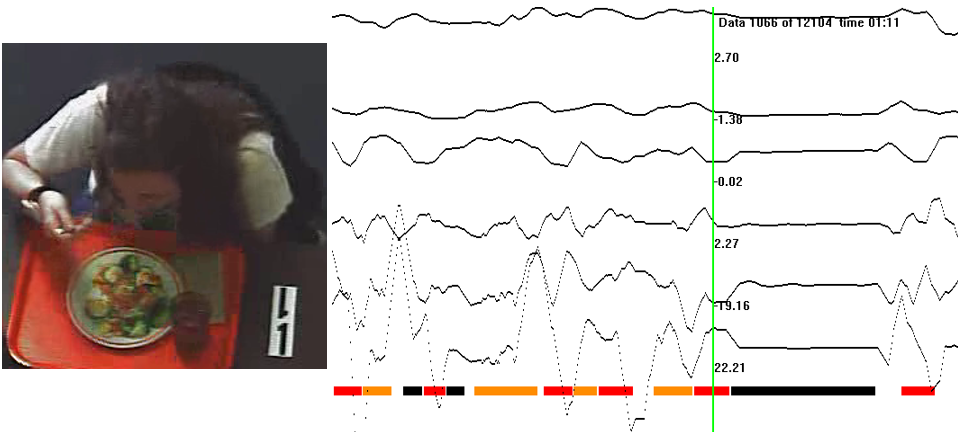}
    \caption{A custom program for gesture labeling. Boxes with different colors indicate gesture types: red = bite, aqua = drink, orange = utensiling, black = rest and grey = other. Sensor data from top to bottom: X, Y, Z, yaw, pitch, roll.}
	\label{fig:label-tool}
\end{figure*}

Figure \ref{fig:label-tool} shows a custom program developed to facilitate labeling and examples of labeled gestures.
The left panel displays the video while the right panel shows the synchronized wrist motion tracking data. 
Top to bottom on the right panel shows 6 axes of motion (accelerometer x, y and z; gyroscope yaw, pitch and roll).
The vertical green line indicates the time currently displayed in the video. 
Boxes laid over the seventh line indicate periods of time labeled as gestures (for example, red = bite). 
Unlabeled segments with duration longer than 4 seconds are considered as other, unlabeled segments shorter than 4 seconds are considered transitions between gestures and are ignored \cite{Ramos15}.

\subsection{Data Preprocessing}
\label{subsec:preprocess}

Let the data set be comprised of $d$ eating sessions, where each session is a recording of wrist motion during contiguous consumption.
For example, a meal might be divided into multiple eating sessions (appetizer, entree, dessert) separated by periods of non-eating that are not recorded.
The wrist motion data is defined as $R_{t}^{d} = [x_{t}^{d}, y_{t}^{d}, z_{t}^{d}, \alpha_{t}^{d},  \beta_{t}^{d},  \gamma_{t}^{d}]$, where $d$ indicates the eating session, $t$ represents the time index, $x, y, z$ are accelerometer sensor readings and $\alpha, \beta, \gamma$ are gyroscope sensor readings.
All the data were smoothed using a Gaussian-weighted window of width 1 s and standard deviation of $\frac{2}{3}$ s:
\begin{equation}
\tilde{R}_{t}^{d} = \sum\limits_{i=-N}^0 R_{t+i}^{d} \cfrac{exp \left(\frac{-t^{2}}{2\sigma^{2}} \right)}{\sum\limits_{x=0}^N exp \left(\cfrac{-(x-N)^{2}}{2\sigma^2} \right)}
\label{eq:smooth}
\end{equation}

\subsection{Hidden Markov models}

\begin{figure*}
	\centering
	\includegraphics[width=0.95\textwidth]{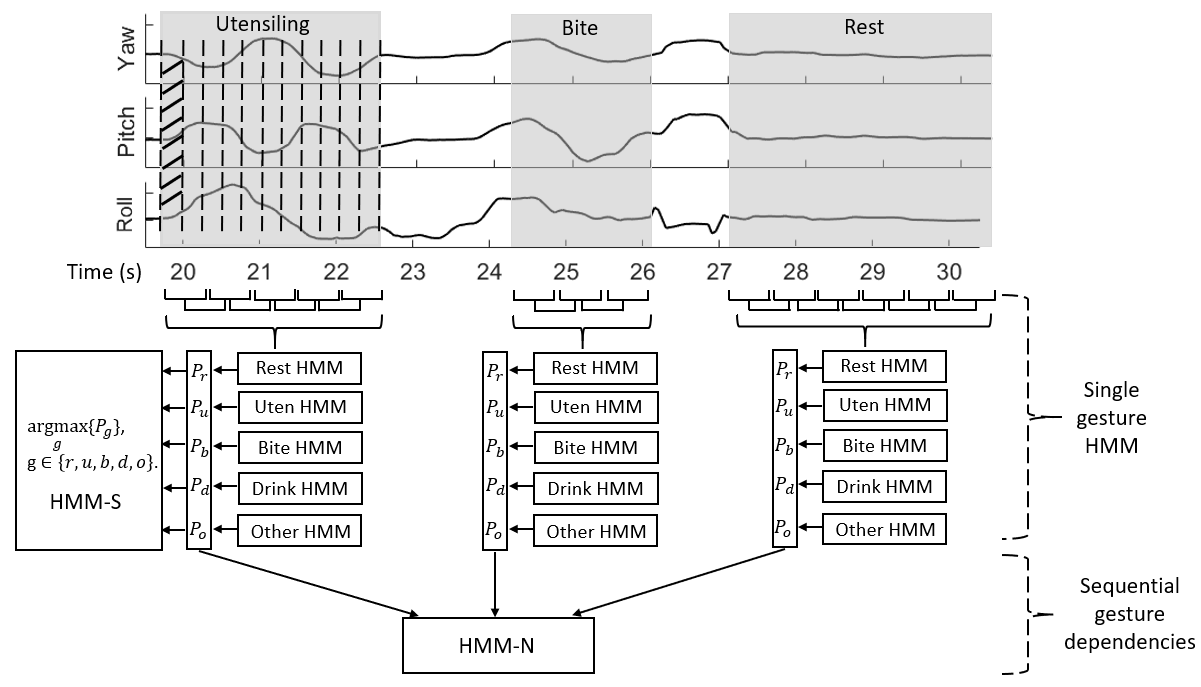}
\caption{Architecture for single gesture HMM-S and gesture-to-gesture HMM-N.
Examples of three manually segmented gestures are displayed.
In HMM-S, the observables are a sequence of features computed from the
raw sensor data (only gyroscope signals are shown for brevity) in
sliding windows, each with 50\% overlap denoted by the shaded area.
Each gesture type (rest, uten., etc.) is recognized using a different HMM.
For each input sequence, the HMM with the maximum logarithmic probability
determines the gesture type.
Gesture sequence recognition uses the set of logarithmic probabilities
as observables for HMM-N, in which each state represents a sequence of
N gestures.}
	\label{fig:architecture}
\end{figure*}

Figure \ref{fig:architecture} illustrates the architecture of our HMMs.
There are two levels, HMM-S and HMM-N.
The first level observes the motion subcomponents
of a single gesture and classifies the motion sequence according to which
gesture it most resembles.
The second level observes the probable identities of a preceding set of
gestures and classifies the current gesture according to which gesture
sequence it most resembles.
The details of each level of HMM are as follows.

\subsubsection{Single Gesture HMM-S}
\label{subsec:baseline}

We use HMM-S to model a single gesture as a sequence of sub-gestures
with each sub-gesture represented by a state \cite{Ramos15}.
For example, the action of taking a bite may consist of raising food
towards the mouth, ingestion, and the return of the wrist to a rest position. 
This sequence is modeled through a state sequence where each state
models part of the motion pattern.

We use the notation $\lambda = (\pi, A, B)$ for each HMM,
where $A$, $B$ and $\pi$ are the state transition matrix,
emission probability and initial state distribution, respectively.
We denote individual states as $S = \{s_{1}, s_{2}, ..., s_{N}\}$,
the state at time $t$ as $q_{t}$, and a
state sequence as $Q = \{q_{1}q_{2}, ..., q_{T}\}$.
The initial state distribution $\pi$ is computed as:
\begin{equation}
\pi_{i} = P(q_{1} = s_{i}), \quad 1 \le i\le N.
\end{equation}
The state transition matrix $A$ is computed as:
\begin{equation}
a_{ij} = P(q_{t+1} = s_{j} | q_{t} = s_{i}), \quad 1 \le i, j \le N, 1 \le t \le T.
\end{equation}

The observables $O$ for each gesture $g$ are calculated
using two windows $w_{1}$ and $w_{2}$, where $w_{1}$ is the length of time in which features are calculated and $w_{2}$ is the step in time between feature calculations.
Formally, we calculate features as in Equations \ref{eq:mu}-\ref{eq:slope}:

\begin{equation}
\mu_{g, t} = \frac{1}{w_{1}} \sum\limits_{i=0}^{w_{1}} \tilde{R}_{g, t+i}
\label{eq:mu}
\end{equation}

\begin{equation}
\sigma_{g, t} = \sqrt{\frac{1}{w_{1}-1} \sum\limits_{i=0}^{w_{1}} (\tilde{R}_{g, t+i} - \mu_{g, \tilde{R},t})^{2}}
\label{eq:sigma}
\end{equation}

\begin{equation}
s_{g, t} = \frac{(\tilde{R}_{g, t+w_{1}} - \tilde{R}_{g, t})}{w_{1}}
\label{eq:slope}
\end{equation}
where $\tilde{R}_{g, t}$ is the smoothed sensor reading of gesture g at time t, and $\mu_{g, t}$, $\sigma_{g, t}$, $s_{g, t}$ are the average, standard deviation and slope.
In each window $w_{1}$, this provides 18 features $o_{g, t} = [\mu_{x_{g, t}}, \mu_{y_{g, t}}, ..., \sigma_{x_{g, t}}, \sigma_{y_{g, t}}, ..., s_{x_{g, t}}, s_{y_{g, t}}, ..., s_{\gamma_{g, t}}]$.
The features for each gesture g can be represented as $O_{g}$ = [$o_{g, t}, o_{g, t+w_{2}}, ..., o_{g, t+l\times w_{2}}$], where $l$ depends on the duration of each gesture.

We select $w_{1}$ and $w_{2}$ to act as a sliding window with overlap.
The classic approach in speech recognition is to use a 0.5 second window with 50\% overlap \cite{lee2012automatic}.
In this paper, since our data is collected at 15 Hz, we use a 9 sample window (0.6 s) with 4 sample overlap (0.3 s). 
We perform a z-score independently for each of the 18 features to prevent skew towards large valued features.

Gaussian mixture models (GMMs) are used to describe the emission
probabilities $B$, as shown in Equation \ref{eq:emission1},
where M is the number of Gaussians.
\begin{equation}
\begin{split}
B = P(O|Q) = \sum\limits_{i=1}^M c_{i}N(O;\mu_{i},\Sigma_{i}), \quad \sum\limits_{i=1}^M c_{i} = 1.
\label{eq:emission1}
\end{split}
\end{equation} 
Each Gaussian is defined by three parameters $c_{i}$, $\mu_{i}$, and $\Sigma_{i}$ representing weight, mean and covariance matrix of the $i^{th}$ Gaussian, respectively:
\begin{equation}
N(O;\mu_{i},\Sigma_{i}) = \cfrac{1}{(2\pi)^{D/2}|\Sigma|^{1/2}} e^{-\frac{1}{2}(O-\mu)^{T} \Sigma^{-1} (O-\mu)}
\label{eq:emission2}
\end{equation} 
We assume features are independent and the off-diagonal entries in $\Sigma$ are zero.
The expectation-Maximization algorithm is used to calculate the
emission probabilities modeled by GMMs \cite{hmmtutorial, xuan2001algorithms}.

An HMM toolbox was used to build HMMs \cite{hmmtoolkit}. 
We use an architecture of left-to-right with skip in HMM-S \cite{Shen16},
so $\pi$ is always one for the first state and zero for the other states.
The forward-backward algorithm is used to train an HMM,
in other words to learn the $A$ and $B$ matrices given an observation
sequence $O$.

In HMM-S, five HMMs are built, one for each gesture type.
During recognition, observables $O$ from an unknown gesture are passed
into the HMMs as illustrated in Figure \ref{fig:architecture}.
Each HMM $\lambda_{g}$ computes the likelihood of this particular
observation sequence using the forward algorithm:
\begin{equation}
P(O | \lambda_{g}) = \sum_{Q} P(O, Q | \lambda_{g}) = \sum_{Q} P(O | Q, \lambda_{g})P(Q | \lambda_{g}).
\end{equation}
Each unknown gesture obtains probability scores from each of the five HMMs.
A probability score indicates how well a model matches the gesture. 
Therefore, the model which provides the maximum score determines the
gesture type g $\in$ \{rest, utensiling, bite, drink, other\} as

\begin{equation}
\begin{split}
\hat{g} &= \argmax_{g} \{P_{g}\}
\end{split}
\label{eq:generic-recognition}
\end{equation}

\subsubsection{Sequential Dependent HMM-N}
\label{subsec:sequential-hmm}

We use HMM-N to model a sequence of N previous gestures as context to
improve recognition of the current gesture.
For example, a common pattern is to use utensils to prepare a bite of
food (U), consume the bite of food (B), and then rest hands while
masticating and swallowing (R).
In HMM-N, each state models a sequence of N gestures.
Figure \ref{fig:architecture} illustrates the architecture of HMM-N.
Note that the observables and states in HMM-N are different from those
used for HMM-S.

\begin{figure}
	\centering
	\includegraphics[width=0.4\textwidth]{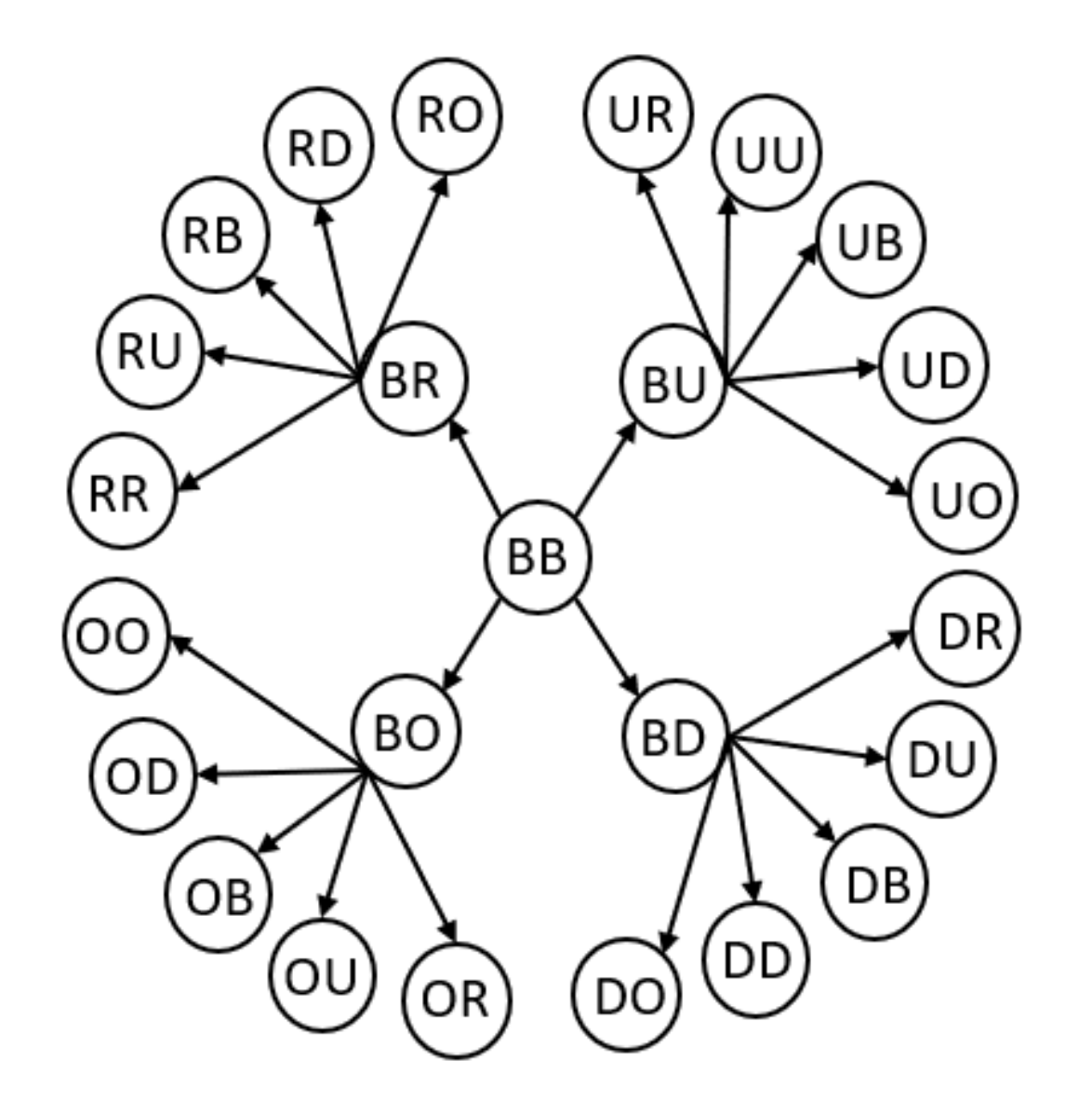}
    \caption{State transitions in HMM-2. For clarity, only the 25 transitions starting from bite are displayed. B = bite, D = drink, R = rest, U = utensiling, O = other.}
	\label{fig:HMM2-state-transition}
\end{figure}

To calculate the state transition matrix, we convert HMM-N to an equivalent first-order HMM \cite{mari1997automatic, Ramos15}.
Figure \ref{fig:HMM2-state-transition} shows a partial example of the equivalent first-order HMM for HMM-2 (for clarity, only state transitions starting
from bite are shown).
Logically, transitions between some states are impossible. 
For example, state BB cannot transition to DB because the former state's most recently recognized gesture is B, which does not match the memory of the latter, which is D.
In total HMM-2 has only $5 \times 25$ possible state transitions.
Formally, each state in HMM-N is $s_{i} = \{g_{1}g_{2}...g_{N}\}$.
The state transition matrix $A$ is calculated as:
\begin{equation}
\begin{split}
a_{ij} &= P(s_{j} = g_{2}g_{3}...g_{N+1}|s_{i} = g_{1}g_{2}...g_{N}) \\
&= \frac{\# \textrm{transitions}\,   \textrm{from}\,   g_{1}g_{2}...g_{N}\,   \textrm{to}\,   g_{2}g_{3}...g_{N+1}+1}{\#\,   g_{1}g_{2}...g_{N}\,     \textrm{gesture sequences}+|S|}
\label{eq:high-transition}
\end{split}
\end{equation}
where $|S|$ indicates the number of possible state transitions.
Laplace smoothing (+1 in numerator, +$|S|$ in denominator)
is used to avoid values of zero in the state transition
matrix (cases in which a sequence does not appear in the training data)
\cite{schutze2008introduction}.

The initial state distribution $\pi$ is calculated as:
\begin{equation}
\begin{split}
\pi_{i} &= P(s_{i} = g_{1}g_{2}...g_{N}) \\
&= \frac{\#\,   g_{1}g_{2}...g_{N}\,       \textrm{gesture sequences}+1}{\#\,    \textrm{N-gesture}\,   \textrm{sequences}+|S|} \\
\label{eq:high-prior} \\
\end{split}
\end{equation}

The observables $O$ of each gesture are the five probability scores from HMM-S.
In HMM-N, we make an assumption that the observables are only dependent on the most recent gesture.
For example, observables from gesture D, gesture sequence UD and UUD are the same (they are all immediate observations of D).
The emission probabilities $B$ are calculated as:
\begin{equation}
\begin{split}
B &= P(O | s_{i} = g_{1}g_{2}...g_{N}) \\
&= P(O | g_{N}) = \sum\limits_{i=1}^{M} c_{i}N(O;\mu_{i},\Sigma_{i}).
\label{eq:high-emission}
\end{split}
\end{equation}
We follow the same parameter setting in \cite{Ramos15} to use 7 Gaussians.

During training, $A$ and $\pi$ are calculated using Equations \ref{eq:high-transition}-\ref{eq:high-prior} and
$B$ is learned using Equation \ref{eq:high-emission}.
Recognition is defined as finding the most likely state sequence $Q = \{q_{1}, q_{2}, ...q_{T}\}$ that explains observables O given model $\lambda = \{A, B, \pi\}$:
\begin{equation}
\begin{split}
Q &= \argmax_{q_{1}, q_{2}, ..., q_{T}} P(Q | O, \lambda)
\label{eq:decoding}
\end{split}
\end{equation}
The Viterbi algorithm \cite{hmmtutorial} is used and the most recent gesture in each $q_{t}$ determines the gesture type for each time step.

\subsection{Model Complexity and Training Data}
\label{subsecsec:method-model-complexity}

To study the amount of motion variability within each gesture type,
we varied the number of states $N$ and the number of Gaussians $M$ for HMM-S.
Specifically we built every combination of HMM-S with $N=3...25$ and
$M=1...7$.
The value for $N$ can be considered to correspond to the number of different
sub-motions expected in a gesture type.
The value for $M$ can be considered to correspond to the number of
observed variations of each expected sub-motion.
During training, we randomly selected 650 gestures of each type
to train the 5 HMMs.
During recognition, we selected another set of 650 gestures per type,
excluding those used in training, to test the accuracy.
Due to the Monte Carlo nature of HMM training,
each model was run 5 times and the average is reported.

To study the effect of the quantity of training data, we varied the
amount of gesture samples used to train each HMM.
The values for $N$ and $M$ were held constant according to the best
values found from the model complexity experiment.
We varied the number of training samples per gesture type from 65 to 650
by randomly selecting from the full data set.
During recognition, the same set of testing data as above was used.
To reduce the variance introduced in the process of random selection of training data, each model was run 30 times and the average is reported.

\section{Results}
\label{sec:results}

The total data set consists of 51,614 manually labeled gestures,
with 14,761 rest, 14,861 utensiling, 18,462 bite, 2,182 drink and 1,348 other. 
This data set, along with a visualization tool, is being made
publicly available at \url{http://cecas.clemson.edu/~ahoover/cafeteria}.

\subsection{HMM-S}
\label{results-hmm-S}
 
\begin{figure}
	\centering
	\includegraphics[width=0.5\textwidth]{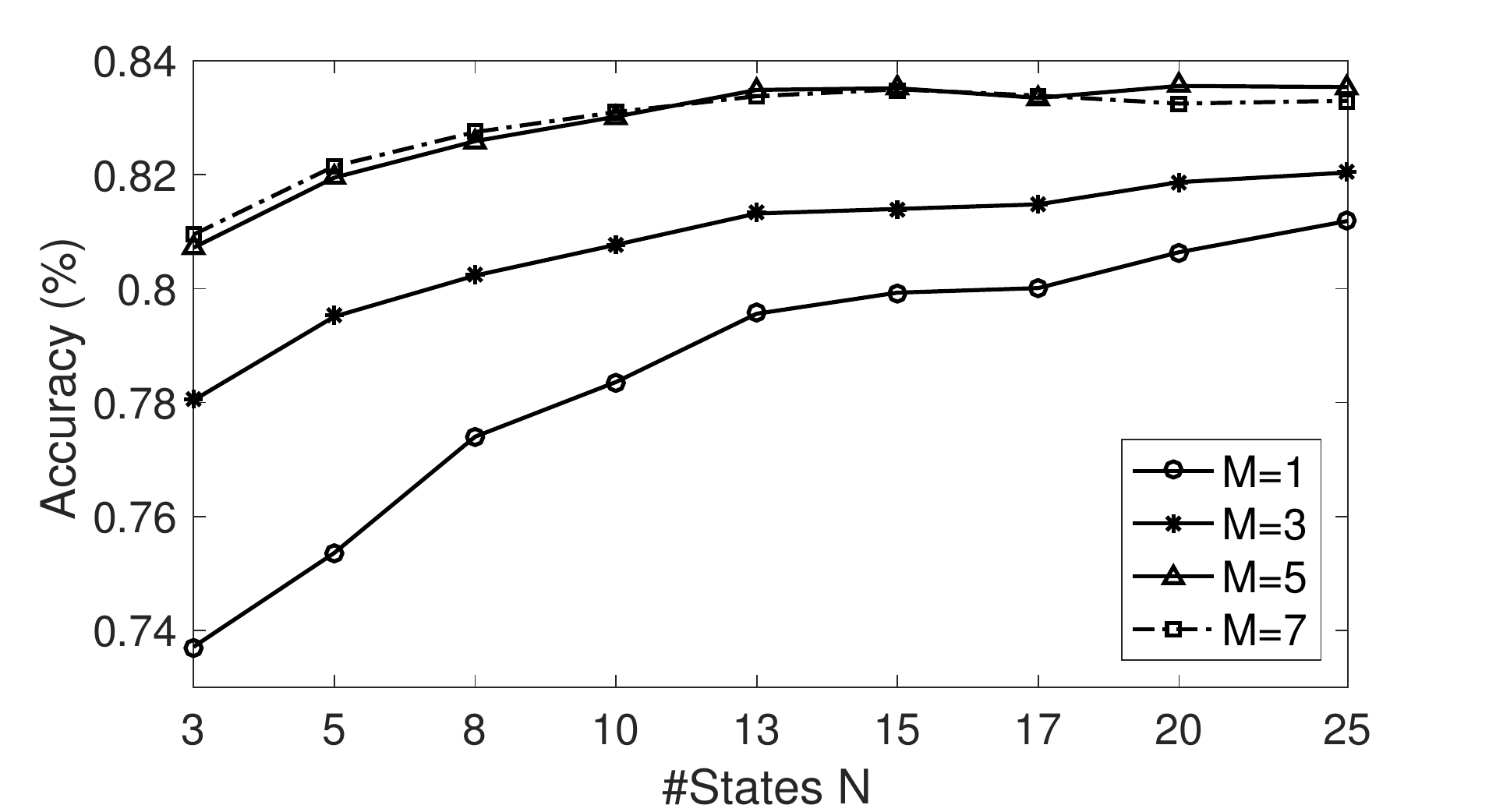}
    \caption{Recognition accuracy with model complexity: the number of states $N$ and mixture components $M$.}
	\label{fig:model-complexity}
\end{figure}

Figure \ref{fig:model-complexity} shows the results for recognition
accuracy vs HMM complexity.
The accuracy plateaus at $M=5$ and $N=13$, indicating that 5 Gaussians and
13 states are needed to capture the motion variability.

\begin{figure}
	\centering
	\includegraphics[width=0.5\textwidth]{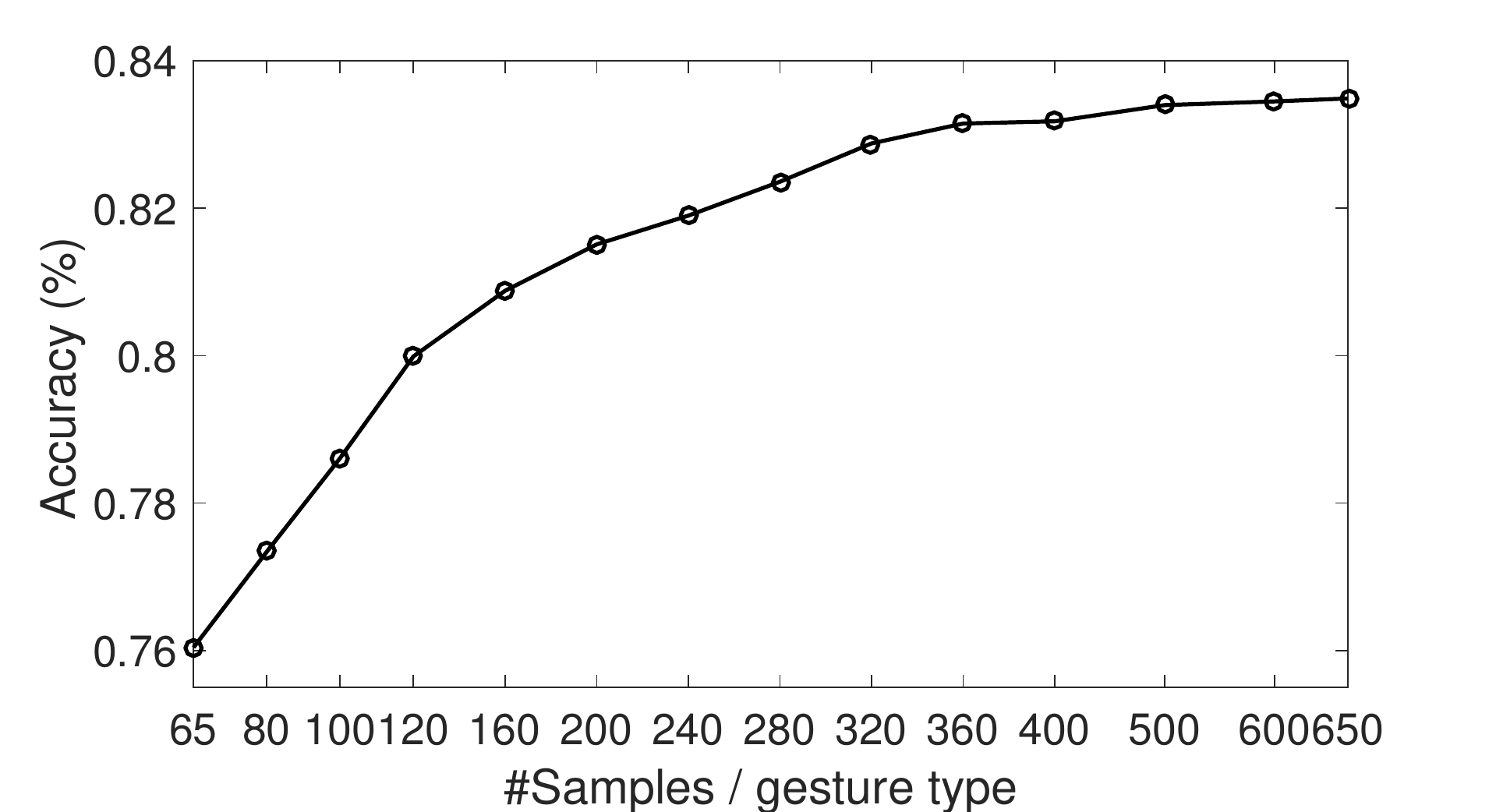}
    \caption{Recognition accuracy with the quantity of training data.}
	\label{fig:impact-training-data}
\end{figure}

Figure \ref{fig:impact-training-data} shows recognition accuracy vs number
of training gestures, with model complexity fixed at $M=5$ and $N=13$.
The accuracy plateaus at 500 training samples per gesture type, indicating
that while 65 training samples per gesture type are the minimum needed
to train HMMs of this complexity, an additional 8\% accuracy is achieved
by training with 500 samples per gesture type.

\subsection{HMM-N}
\label{results-hmm-N}

\begin{figure}
  \centering
  \includegraphics[width=0.5\textwidth]{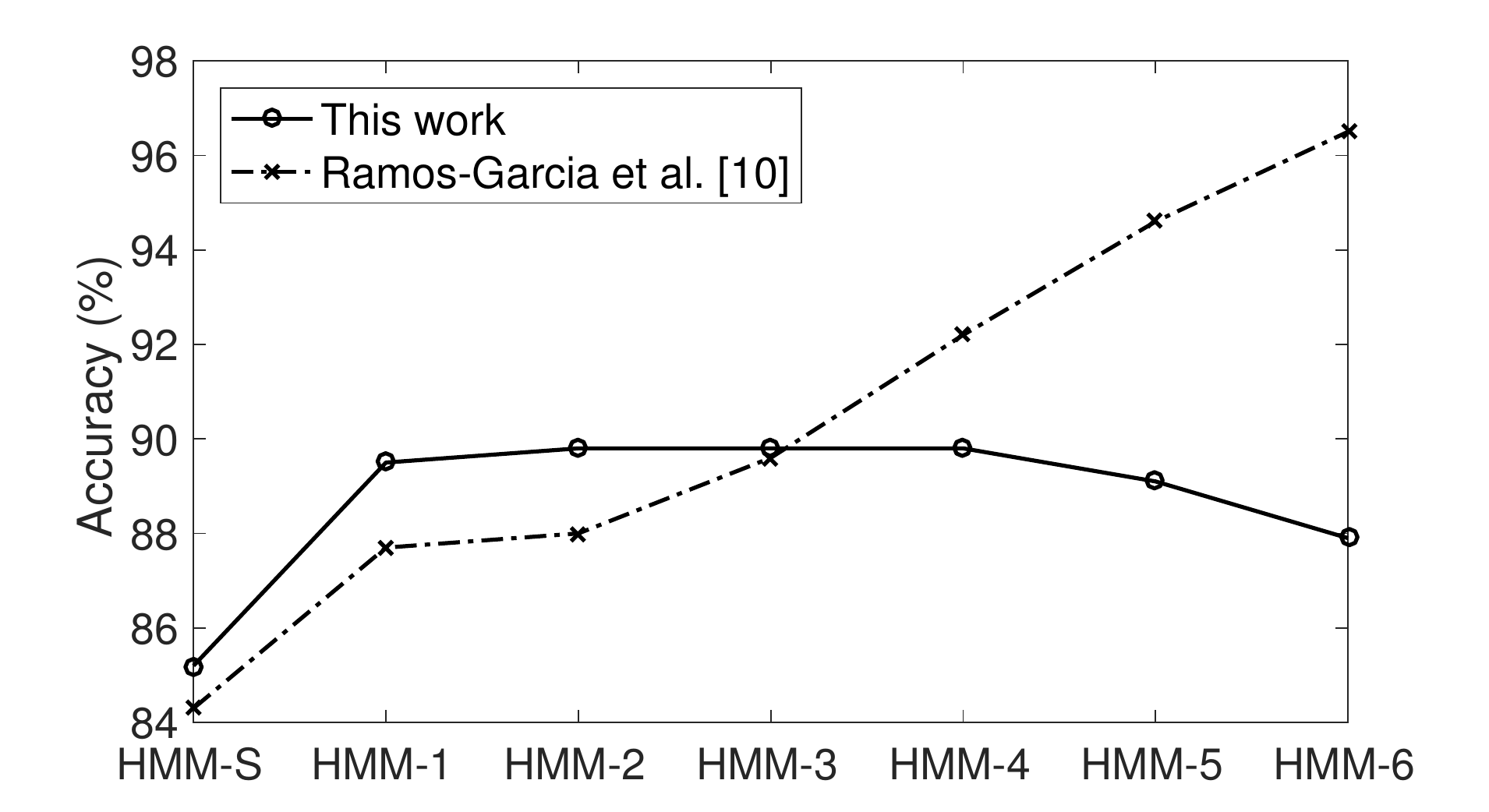}
  \caption{Accuracy of models trained on different amount of data.}
  \label{fig:fullset-vs-subset}
\end{figure}

Figure \ref{fig:fullset-vs-subset} shows the accuracy for HMM-N as
different amounts of previous gestures (S=0, N=1...6) are incorporated
into the classifier.
The figure compares our results that used
51,614 gestures for training vs the same model trained on a smaller data
set of 2,786 gestures from 25 meals \cite{Ramos15}.
The accuracy of HMM-1 is improved by 1.8\% by having more training data,
although part of this can be explained by the simultaneous 0.9\% increase
in accuracy in HMM-S that produces the inputs used for HMM-1.
However, the accuracies for HMM-2 and higher are not improved by more
training data.
One reason is that building HMM-N requires converting high-order HMMs to first order by enumerating every possible combination between current and previous gestures.
This requires a large amount of parameters in states and transition matrices, as shown in Table \ref{table:param-amount}.
In HMM-N, the number of states is $5^{N}$ and possible state transitions is $5 \times 5^{N}$.
Emission probabilities are modeled by mean $\mu$ and covariance $\Sigma$ of GMMs, as shown in Equation \ref{eq:high-emission}. 
Given M mixture components and D-dimensional observables, the amount of emission probability parameters are $5\times D\times M$ for $\mu$ and $\Sigma$, respectively.
Based on the one in ten rule in training models \cite{harrell1984regression}, it is necessary to build models with a quantity of training data at least 10 times the number of parameters.
Therefore, the inadequate training data in \cite{Ramos15} caused overfitting for models from HMM-3 to HMM-6.
It is worth noting that even with the large data set of 51,614 gestures, we still do not have adequate data to train HMM-5 and HMM-6.
Another reason is that the transition matrix of testing data in \cite{Ramos15}
was included during training to avoid cases where gesture sequences in
the training data did not exist in the testing data,
but this biased the results.
In this work, we used Laplace smoothing instead.

\begin{table}
\centering
\begin{tabular}{| p{1.1cm} | p{0.8cm} | p{0.8cm} | p{0.8cm} | p{0.8cm} | p{0.8cm} | p{0.8cm} |} \hline 
$\text{\#Params.}$& HMM-1 & HMM-2 & HMM-3 & HMM-4 & HMM-5 & HMM-6 \\ \hline 
Prior & 5 & 25 & 125 & 625  & 3,125 & 15,625 \\
Transition & 25 & 125 & 625 & 3,125 & $15,625$ & 78,125 \\
Emission & 350 & 350 & 350 & 350 & 350 & 350 \\ \hline
Total & 380 & 500 & 1,100 & 4,100 & 19,100 & 94,100 \\ \hline
\end{tabular}
\caption{\#Parameters in HMM-$N$. Note: observable is 5-dimensional vector and 7 GMMs are used.}
\label{table:param-amount}
\end{table}

Finally,
Table \ref{tab:history-hmm-overall-table} summarizes the accuracies of
HMM-S and HMM-1 trained and tested on all data using five-fold cross
validation.
Overall, HMM-S achieves 85.2\% accuracy and HMM-1 achieves 89.5\% accuracy.
For each gesture, the improved accuracy for rest, utensiling and bite is 2.2\%, 4.3\%, 8.2\%, respectively.
We observe that drink and other decrease in accuracy from HMM-S to HMM-1,
suggesting that there is not enough sequencing consistency
in a large data set to warrant modeling their sequencing in an HMM.

\begin{table}
\centering
\begin{tabular}{| c | c | c | c | c | c | c | c |} \hline
Model & All   & Rest & Utensiling & Bite & Drink & Other \\ 
& (\%) & (\%) & (\%) & (\% ) & (\%) & (\%) \\ \hline
HMM-S & 85.2 & 86.8 & 86.9 & 83.7 & 96.1 & 52.6 \\
HMM-1 & 89.5 & 89.0 & 91.2 & 91.9 & 91.2 & 52.2 \\ \hline  
\end{tabular}
\caption{Recognition accuracy for HMM-S and HMM-1.}
\label{tab:history-hmm-overall-table}
\end{table}

\section{Discussion}
\label{sec:discussion}

Motivated by works in sign language to vary model complexity and quantity of training data to identify the point of diminishing returns in recognition accuracy, this paper describes a similar effort for the recognition of eating gestures.
Two models were built: HMM-S which models the sequence of actions within a gesture, and HMM-N which models the sequential dependence between gestures.
A total of 51,614 gestures with 5 different gesture types were labeled from 269 subjects eating a single meal in a cafeteria environment.
Sign language HMMs constructed with only 3-4 states and 1-3 Gaussians achieved more than 90\% accuracy \cite{fang2007large, kumar2017coupled}, whereas we found that eating gesture HMMs needed 13 states and 5 Gaussians to achieved 85.2\% accuracy.
For training data, in contrast to sign language in which only 24-60 training samples per word were sufficient, we found that 500 training samples per gesture type were required.
These findings demonstrate that the variability of motion patterns in eating gestures are much larger than the variability in motion patterns in sign language, and that more complex models and more training data are required.
For HMM-N, the effect of model complexity was explored by studying the sequential dependence of N previous gestures to improve recognition of the current gesture.
Results show that accuracy was improved by 4.3\% when one previous gestures was studied as context, but was not improved when additional previous gestures were studied. 
This demonstrates that word/gesture sequencing, commonly used to improve
speech/sign language recognition accuracy, may have less applicability to
eating gesture recognition.

One of the key challenges for eating gesture recognition is to translate models built with laboratory data to models built with free-living data.
Most previous works have trained models with data in a limited amount of subjects, time collected and lab environments \cite{amft2005detection, ye2015automatic, thomaz2017exploring}.
However, a large variability exists in eating gestures during free-living and models that are developed in a controlled setting will potentially be brittle in a natural setting.
For example, people can gesticulate while talking to others or place a phone call while eating, activities which may not happen in a laboratory environment.
Recently, a few studies have begun investigating the differences between
performance in laboratory and real-world environments.
One study \cite{thomaz2015practical}
trained models using 10 hours of data collected in a lab from
20 subjects and tested on 31 hours of data collected in free-living from
7 subjects, compared to 422 hours of data collected in free-living from
1 subject, and found 76\% and 71\% F-scores, respectively.
Another study \cite{mirtchouk2017recognizing} trained models using
59 hours of data collected in the lab from 6 subjects and tested
on 113 hours of data collected in free-living from the same group of
subjects, finding 88\% precision and 87\% recall.
A third study \cite{zhang2018monitoring} detected chewing events
on 122 hours of data collected in free-living from 10 subjects,
finding 79\% recall and 77\% precision, respectively.
However, it is difficult to directly compare these efforts with our work.
First, they all investigated different sensing modalities including
sensing of wrist motion, chewing and swallowing.
Second, these works focus only on eating detection, while our work
recognizes individual activities during eating.

A limitation of this study is that only one meal was collected per subject,
which might not be adequate to capture variability within individuals.
Another limitation is that data was collected only in a cafeteria location, which might not capture the full motion variability of free-living eating behaviors.
In future work we would like to collect free-living data from individuals over a longer duration (a week or more) and in multiple locations.

\section*{Acknowledgment}
We gratefully acknowledge the support of the NIH via grant 1R01HL118181-01A1.

\end{document}